\documentclass[letterpaper]{article} % DO NOT CHANGE THIS
\usepackage{aaai2026}  % DO NOT CHANGE THIS
\usepackage{times}  % DO NOT CHANGE THIS
\usepackage{helvet}  % DO NOT CHANGE THIS
\usepackage{courier}  % DO NOT CHANGE THIS
\usepackage[hyphens]{url}  % DO NOT CHANGE THIS
\usepackage{graphicx} % DO NOT CHANGE THIS
\usepackage{natbib}  % DO NOT CHANGE THIS AND DO NOT ADD ANY OPTIONS TO IT
\usepackage{caption} % DO NOT CHANGE THIS AND DO NOT ADD ANY OPTIONS TO IT
\usepackage{algorithm}
\usepackage{multirow}
\usepackage[table]{xcolor}
\usepackage{booktabs}
\usepackage{amssymb}
\usepackage{subcaption}
 \usepackage{amsmath}
 \usepackage{algpseudocode}
\usepackage{algorithm}
\usepackage{multirow}
\usepackage[table]{xcolor}
\usepackage{booktabs}
\usepackage{amssymb}
\usepackage{subcaption}
 \usepackage{amsmath}
 \usepackage{algpseudocode}
\usepackage{pifont}

\usepackage[most]{tcolorbox}
\usepackage{newfloat}
\usepackage{listings}
\DeclareCaptionStyle{ruled}{labelfont=normalfont,labelsep=colon,strut=off} % DO NOT CHANGE THIS
\floatstyle{ruled}
\newfloat{listing}{tb}{lst}{}
\floatname{listing}{Listing}
\title{Preference-Aware Memory Update for Long-Term LLM Agents}
\author{
    Haoran Sun, %\textsuperscript{\rm 1}, %\thanks{With help from the AAAI Publications Committee.}\\
    Zekun Zhang,
    Shaoning Zeng\\
}
\usepackage{bibentry}
\begin{document}
\maketitle

\begin{abstract}

One of the key factors influencing the reasoning capabilities of LLM-based agents is their ability to leverage long-term memory. Integrating long-term memory mechanisms allows agents to make informed decisions grounded in historical interactions. While recent advances have significantly improved the storage and retrieval components—e.g., by encoding memory into dense vectors for similarity search or organizing memory as structured knowledge graphs—most existing approaches fall short in memory updating. In particular, they lack mechanisms for dynamically refining preference memory representations in response to evolving user behaviors and contexts. To address this gap, we propose a Preference-Aware Memory Update Mechanism (PAMU) that enables dynamic and personalized memory refinement. By integrating sliding window averages (SW) with exponential moving averages (EMA), PAMU constructs a fused preference-aware representation that captures both short-term fluctuations and long-term user tendencies. We conduct experiments on five task scenarios of the LoCoMo dataset, and the results show that our mechanism can significantly improve the output quality of LLM in five baselines, validating its effectiveness in long-term conversations.

%The capabilities of large language models (LLMs) rely on pre-trained fixed model weights, but at the same time, the boundaries of LLMs' capabilities are also limited to a certain extent by pre-training. Utilizing memory mechanisms or preference optimization frameworks to align LLM outputs more closely with human preferences is a common strategy. However, these approaches only address user preferences while neglecting the enhancement of the LLM’s own domain understanding abilities—they optimize the emotional intelligence (EQ) of the LLM but overlook its intelligence quotient (IQ), thus limiting the LLM's ability to self-evolution. To address this, we propose a Dual-Phase Self-Evolution (DPSE) Framework to enhance alignment with user preferences and domain knowledge. DPSE first employs a Censor module to filter dialogue samples based on Customer Satisfaction Scores and domain attention. These samples guide data generation for supervised fine-tuning and preference optimization. A subsequent two-stage self-evolution strategy further refines model alignment. We conduct comparative experiments against supervised fine-tuning, preference optimization, and memory-based methods. The results demonstrate that DPSE consistently outperforms these baselines, validating its overall effectiveness.

\end{abstract}


\section{Introduction}

Large Language Model (LLM) agents exhibit strong autonomous decision-making capabilities across a wide range of tasks, particularly excelling in open-domain question answering \cite{LLMsurvey1,agentsurvey,deepseek-r1}. In long-term dialogue scenarios, effective reasoning and decision-making often require integrating past interactions, making internal memory mechanisms essential \cite{memorysurvey,memorysurvey1}. These mechanisms aim to emulate human-like cognitive memory by retaining prior conversational context, enabling the agent to retrieve relevant information and generate context-aware, personalized responses. The design and adaptation of such memory systems are thus critical to the agent’s performance in complex, temporally extended tasks \cite{agentsurvey2,agentsurvey3,datasetagent,INoT}.

The most basic memory approach concatenates prior dialogues with the current prompt, but this method is constrained by the LLM's finite context window, limiting its effectiveness in prolonged interactions \cite{agentsurvey4,LLMsurvey2}. To address this, recent studies have explored more sophisticated architectures: MemoryBank \cite{memorybank} encodes past information into dense vectors and retrieves memories via similarity search ; MemGPT \cite{memgpt} introduces a hierarchical OS-inspired memory system that combines limited-context attention with external memory storage, yet suffers from a trade-off between retrieval accuracy and efficiency; MemInsight \cite{meminsight} enhances memory representation by autonomously extracting structured key-value attributes; and A-MEM \cite{Amem}, inspired by the Zettelkasten method, dynamically constructs evolving knowledge graphs for self-organizing memory.

Despite these advances, existing systems predominantly focus on memory storage and retrieval, often overlooking a crucial aspect: how to adaptively and continuously update memory in response to evolving user behavior during long-term interactions \cite{LLMsurvey5,LLMsurvey6}. In real-world deployment, users are non-stationary—their intents, preferences, and goals shift over time. Without dynamic memory updating, agents risk relying on outdated or misaligned information, leading to degraded performance and user trust.

To bridge this gap, we propose a Preference-Aware Memory Update Mechanism that enables LLMs to perceive, adapt to, and respond in alignment with evolving user preferences. At its core is a novel Preference Change Perception Module, which combines a sliding window average and an exponential moving average (EMA) to construct a dual-perspective user preference representation—capturing short-term behavioral shifts while robustly modeling long-term trends. We further introduce a formalized change detection signal, triggered by the deviation between short- and long-term estimates, to guide when and how memory updates should occur. This allows for interpretable and controllable adaptation in response to preference drift. Notably, our mechanism is highly modular and model-agnostic: it requires no fine-tuning or architectural modification and can be seamlessly integrated into existing memory-augmented LLM frameworks.

\section{Related Work}

\begin{figure*}[t]
    \centering
    \includegraphics[width=1\linewidth]{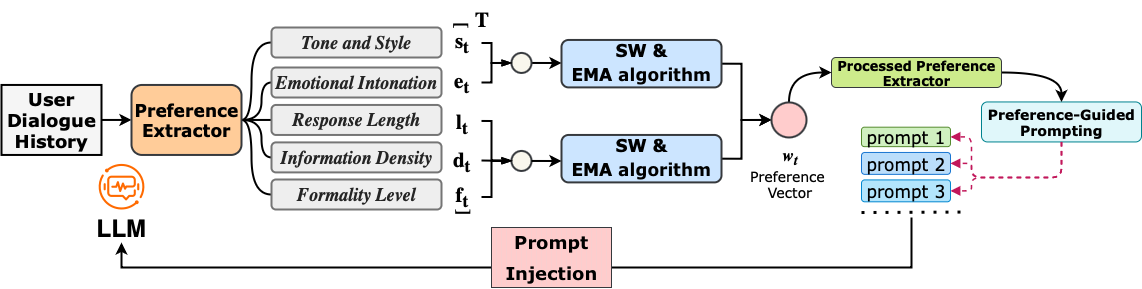}
    \caption{\textbf{Illustration of PAMU method. }PAMU extracts user preferences from dialogue, models short- and long-term trends via SW and EMA, detects preference shifts, and updates the prompt to guide personalized generation.}
    \label{fig:framework}
\end{figure*}

To enhance the long-term reasoning capabilities of LLM agents, various memory systems have been proposed. ReadAgent \cite{readingagent} segments and compresses documents into key-point memories for retrieval-augmented reading comprehension. MemGPT \cite{memgpt} uses OS-inspired virtual memory management, combining hierarchical memory with external storage via dynamic function calls. SCM \cite{SCM} enables agents to autonomously decide when and how to access memory through a controller-stream-agent framework. MemoryBank \cite{memorybank}, grounded in the Ebbinghaus forgetting curve, supports memory storage, retrieval, and update for user-aware personalization. A-MEM \cite{Amem}, inspired by Zettelkasten, organizes memory as evolving, self-linked knowledge notes. MemInsight enhances memory representation by extracting structured attributes for more accurate semantic retrieval \cite{meminsight}.

While these approaches have advanced memory modeling in LLMs—especially in storage, retrieval, and organization—they largely assume static user behavior. In practice, user preferences and goals evolve dynamically. However, existing systems lack mechanisms to adaptively track and update memory in response to such changes. This highlights a critical gap: the need for a dynamic, preference-aware memory update mechanism that supports long-term personalization in LLM agents.

\section{Methodology}
In this section, we introduce our Preference-Aware Memory Update (PAMU) mechanism.
%This section introduces a user preference–oriented memory update mechanism that enables LLMs to remember, understand, and adaptively utilize user preferences while dynamically responding to changes in user behavior. The core mechanism is a Preference Shift Detector, which combines a sliding window with exponential moving average (EMA) to adaptively capture both subtle drifts and abrupt shifts in user preferences, allowing the model to adjust its behavior accordingly.

\subsection{Preference Extractor}

The system constructs a user preference vector $\mathbf{P} = \{p_1, p_2, \dots, p_D\}$ by extracting multidimensional preference signals from multi-turn interactions between the user and the model. Each dimension $p_d$ represents a specific user preference type, such as tone style, response length, emotional tone, information density, and degree of formality.
After each dialogue turn, the system updates the preference vector by analyzing user feedback and linguistic features. Specifically:
\begin{itemize}
    \item \textbf{Tone Style.} A RoBERTa encoder with a multi-class classification head is employed to analyze the stylistic features of user utterances. The model produces a probability distribution over predefined tone categories. The category with the highest probability and its score are concatenated into a tuple to represent the tone dimension.

    \item  \textbf{Response Length.} This is measured by the number of tokens generated by the model. The average response length over the past K turns is computed and normalized to the [0, 1] range to form the length dimension.

    \item \textbf{Emotional Tone.} An emotion classification model identifies the dominant emotional categories from both user and assistant utterances. A probability vector over predefined emotional classes is extracted, and the class with the highest probability is used, along with its score, to represent the emotional tone dimension.

\item \textbf{Information Density. }The system leverages an OpenIE model to extract structured (subject, predicate, object) triples from the assistant’s responses. Each triple is treated as an atomic information unit. The number of extracted triples per turn is treated as the count of knowledge points. The information density $ID_t$ of the response at turn t is defined as:
\begin{equation}
ID_t = \frac{K_t}{L_t}
\end{equation}

Among them, \( K_t \) represents the number of triples sampled in the t-th round, and \( L_t \) represents the total number of words in the response of that round. This ratio measures the average amount of information carried by each word, reflecting the compactness of language use and the degree of knowledge density.

\item \textbf{Degree of Formality.} A pretrained formality classification model is employed to evaluate the assistant’s response, yielding a normalized formality score within the range [0, 1], where 0 indicates fully colloquial (spoken) language and 1 denotes fully formal (written) language. This score is directly used as the value for the formality dimension.

\end{itemize}

\begin{algorithm}[ht]
\caption{Preference-Aware Dialogue Generation}
\label{alg:preference-aware}
\begin{algorithmic}[1]
\Procedure{GenerateResponse}{$H_t$, $x_t$} \Comment{History and current user input}
    \State $p_t \gets \Call{ExtractPreferences}{H_t, x_t}$ \Comment{$p_t = (s_t, l_t, e_t, d_t, f_t)$}
    \ForAll{$d \in \{$tone, length, emotion, density, formality$\}$}
        \State $SW_t[d] \gets \text{Mean}(p_{t-W+1:t}[d])$ \Comment{Sliding window average}
        \State $EMA_t[d] \gets \beta \cdot EMA_{t-1}[d] + (1-\beta) \cdot p_t[d]$
        \State $w_t[d] \gets \lambda \cdot SW_t[d] + (1-\lambda) \cdot EMA_t[d]$
    \EndFor
    \State $desc \gets \Call{FormatPreference}{w_t}$ \Comment{Natural language preference prompt}
    \State $prompt \gets \texttt{"Respond in style: "} + desc + \texttt{"\textbackslash n"} + x_t$
    \State $y_t \gets \text{LLM.generate}(prompt)$
    \State \Return $y_t$
\EndProcedure

\Function{ExtractPreferences}{$H_t$, $x_t$}
    \State $s_t \gets \text{ToneClassifier}(x_t, H_t)$ \Comment{Categorical: RoBERTa-based}
    \State $l_t \gets \text{Normalize}(\text{MeanLength}(r_{t-K:t-1}))$
    \State $e_t \gets \text{EmotionAnalyzer}(x_t, H_t)$
    \State $d_t \gets \text{InfoDensity}(r_{t-1})$ \Comment{Triple/token ratio}
    \State $f_t \gets \text{FormalityDetector}(x_t)$
    \State \Return $(s_t, l_t, e_t, d_t, f_t)$
\EndFunction

%\Function{InfoDensity}{$r$}
   % \If{$r = \emptyset$}
     %   \State \Return $0.5$
   % \Else
    %    \State \Return $\text{TripleCount}(r) / \text{TokenCount}(r)$
   % \EndIf
%\EndFunction

\Function{FormatPreference}{$w_t$}
    \State \Return \texttt{[Tone: }$\text{Label}(w_t[\text{tone}])$\texttt{, Emotion: }$\text{Label}(w_t[\text{emotion}])$\texttt{,}
    \State \hskip1.5em \texttt{Density: }$\text{Quantize}(w_t[\text{density}])$\texttt{, Length: }$\text{Quantize}(w_t[\text{length}])$\texttt{,}
    \State \hskip1.5em \texttt{Formality: }$\text{Quantize}(w_t[\text{formality}])$\texttt{]}
\EndFunction
\end{algorithmic}
\end{algorithm}

\noindent Accordingly, for each dialogue turn, the system extracts a five-dimensional user preference vector:

\begin{equation}
\mathbf{p}_t = (s_t, l_t, e_t, d_t, f_t)
\end{equation}

\noindent Here, $s_t$ and $e_t$ denote tuples containing the predicted category index and its probability for tone style and emotional tone, respectively; $l_t$, $d_t$, and $f_t$ are normalized scalar values representing response length, information density, and formality.
This vector is then fed into the Preference Shift Detector, which models the temporal dynamics of user preferences using a combination of a sliding window mechanism and Exponential Moving Average (EMA). This enables the system to detect both gradual drifts and abrupt shifts in preferences, and to determine whether the model’s response strategy requires adaptation or fine-tuning to better align with evolving user intent.

\subsection{Preference Change Perception Mechanism}

Following the extraction of multi-dimensional user preference vectors, a Preference Dynamics Perception Module is employed to model behavioral shifts and enable personalized response adaptation. This module integrates Sliding Window (SW) averaging with Exponential Moving Average (EMA) to continuously update preference estimates at each dialogue turn, thereby guiding the response generator toward controlled, user-aligned outputs.

Specifically, we uniformly represent user preference vectors in the form of:

\begin{equation}
\boldsymbol{p}_t = \left[ p_t^{(1)}, p_t^{(2)}, \dots, p_t^{(D)} \right]
\end{equation}

\noindent Among them, D represents the number of preference dimensions. Each dimensional preference value \(p_t^{(d)}\) may be a continuous variable (such as response length, information density, formality level) or a categorical variable (such as tone style, emotional intonation). For categorical variables, we use the tuple \((c_t^{(d)}, q_t^{(d)})\) to represent, where \(c_t^{(d)}\) is the category index and \(q_t^{(d)}\) is the categorical probability distribution.

\subsubsection{Dynamic Modeling of Continuous Preference Dimensions. }For continuous preference dimensions (length, information density, and Degree of formalization), we define a sliding window of length W to calculate the sliding average preference value at the current time t.

\begin{equation}
\mathrm{SW}_t^{(d)} = \frac{1}{W} \sum_{i=t-W+1}^{t} p_i^{(d)}
\end{equation}

\noindent Among them, $\mathrm{SW}_t^{(d)}$ is the sliding window average of the preference in the $d$-th dimension at time $t$; $W$ is the sliding window length (the number of historical rounds used to calculate the average); $p_i^{(d)}$ represents the preference value in the $d$-th dimension of the $i$-th round; $\sum_{i=t-W+1}^{t}$ denotes the cumulative operation on the preference values from the $(t-W+1)$-th round to the $t$-th round within the window.

Meanwhile, Exponential Moving Average (EMA) is introduced to enhance the memory capacity for long-term trends. Let EMAt(d) denote the exponential average of preference dimension d at time t, then its update formula is:

\begin{equation}
\text{EMA}_t^{(d)} = \beta \cdot \text{EMA}_{t-1}^{(d)} + (1 - \beta) \cdot p_t^{(d)}
\end{equation}

\noindent Among them, $\beta \in (0, 1)$ is the decay coefficient, which controls the degree of influence of historical preferences on the current estimate. $\text{SW}$ is more sensitive to recent preference changes, while $\text{EMA}$ is used to slowly track long-term trends.

After the combination of the two, the fused perception vector is defined as:

\begin{equation}
\hat{w}_t^{(d)} = \lambda \cdot \text{SW}_t^{(d)} + (1 - \lambda) \cdot \text{EMA}_t^{(d)}
\end{equation}

\noindent Among them, $\lambda \in [0, 1]$ controls the weight proportion of the sliding window and exponential average. This fusion strategy can flexibly adapt to the fast-changing and slow-changing characteristics in user preferences.

\subsubsection{Dynamic Modeling of Categorical Preference Dimensions. }For categorical dimensions (tone style and emotional intonation), we represent the preference of each round as $(c_t^{(d)}, q_t^{(d)})$, which is the currently most likely category and its corresponding probability distribution. We perform sliding average and exponential average on the category probability distribution vectors respectively:

\begin{equation}
\mathrm{SW}_t^{(d)} = \frac{1}{W} \sum_{i=t-W+1}^{t} q_i^{(d)}
\end{equation}

\begin{equation}
\mathrm{EMA}_t^{(d)} = \beta \cdot \mathrm{EMA}_{t-1}^{(d)} + (1 - \beta) \cdot q_t^{(d)}
\end{equation}

\noindent The fused category probability perception vector is:

\begin{equation}
\hat{w}_t^{(d)} = \lambda \cdot \mathrm{SW}_t^{(d)} + (1 - \lambda) \cdot \mathrm{EMA}_t^{(d)}
\end{equation}

\noindent Select the category with the highest probability as the control label to be used during generation at the current time:

\begin{equation}
c_t^{(d)} = \arg\max_j \hat{w}_t^{(d)}[j],
\end{equation}

\noindent where \( j \) is the category index.

\subsection{Preference-Guided Prompting}

To enable personalized generation, we explicitly inject the fused user preference vector $\mathbf{w}_t$ into a structured natural language prompt. This guides the LLM to produce outputs aligned with the user’s desired style and attributes, without modifying the model architecture or decoder—achieving flexible behavior control purely via prompt engineering.

Compared to fine-tuning-based implicit modeling, this approach is more efficient, interpretable, and adaptable at inference time, avoiding issues like catastrophic forgetting and supporting real-time preference updates in multi-user or multi-domain settings.

Concretely, $\mathbf{w}_t$ is converted into a textual instruction embedded in the prompt, e.g., “Please answer the following question in the style of: [Tone: humorous], [Emotion: relaxed], [Information density: moderate], [Length: brief].”

Each preference in the prompt is derived from the current dialogue turn, using a fusion of sliding window averaging and exponential moving average (EMA) to smooth short-term fluctuations. The formatting of different preference types is as follows:

\begin{itemize}
    \item \textbf{Categorical dimensions} (e.g., tone style, emotional tone) represented as tuples (c, p), where c is the index of the most probable category and p its confidence score. The selected label $c^*$ is verbalized into descriptors such as “humorous,” “serious,” or “gentle” for prompt inclusion.
    
    \item \textbf{Continuous dimensions} (e.g., response length, information density, formality) maintained as scalar values, discretized into predefined intervals and mapped to interpretable semantic tags (e.g., “brief,” “detailed”) to enhance the model’s understanding of intensity and alignment strength.
\end{itemize}

\noindent For the information density value $d \in [0, 1]$, we define a discretization function that maps continuous preference scores into interpretable semantic tags:

\begin{equation}
\text{Label}(d) =  \begin{cases}  \text{Sparse}, & d \in [0, 0.33) \\ \text{Moderate}, & d \in [0.33, 0.66) \\ \text{Dense}, & d \in [0.66, 1] \end{cases}
\end{equation}

\noindent This mapping strategy is applied uniformly to all continuous preference dimensions (e.g., response length, information density, formality). By concatenating the resulting descriptors across dimensions, a complete structured control prompt can be automatically constructed.
Such an explicit prompting mechanism enables the preference vector to function not only as a soft controller for generation, but also as an interpretable interface for user-aligned output control. Owing to its model-agnostic nature, this mechanism is highly extensible and applicable to a wide range of downstream tasks, including multi-turn dialogue generation, personalized question answering, and preference-aware memory systems.

\subsection{Motivation and Basis}

In long-term human-computer interaction scenarios, user behavior exhibits strong non-stationarity. Users’ tone styles, emotional states, information density requirements, and degrees of formality often undergo gradual evolution or abrupt changes due to factors such as task context, personal emotions, and interaction stages. Although existing memory mechanisms have made progress in information storage and retrieval, they generally rely on a core assumption: that user preferences are stable or uniformly distributed over time. This static assumption may lead the model to generate responses based on outdated preferences, reducing dialogue consistency and user satisfaction. Therefore, our memory update mechanism must possess sensitivity and behavioral interpretability.

In time-series modeling, Sliding Window Average and Exponential Moving Average (EMA) are two commonly used but complementary techniques. Sliding Window Average is sensitive to recent changes and is suitable for capturing short-term preference fluctuations, while EMA focuses on long-term trends through exponential decay, filtering out local noise and modeling inertial behavior. Thus, we propose to integrate the two, constructing a preference perception vector that is both responsive and stable, allowing the model to balance its response style between short-term personalization and long-term consistency:

\begin{equation}
\hat{w}_t^{(d)} = \lambda \cdot \text{SW}_t^{(d)} + (1 - \lambda) \cdot \text{EMA}_t^{(d)}
\end{equation}

\noindent Where, $\text{SW}_t^{(d)} = \frac{1}{W} \sum_{i = t - W + 1}^{t} p_i^{(d)}$ represents the recent average of preferences; $\text{EMA}_t^{(d)} = \beta \cdot \text{EMA}_{t-1}^{(d)} + (1 - \beta) \cdot p_t^{(d)}$ represents the smoothed trend of historical preferences. $\lambda \in [0,1]$ controls the degree of attention to short-term changes, and $\beta \in (0,1)$ controls the memory depth of long-term trends. The mechanism is theoretically justified from the following three perspectives:

\subsubsection{1. Bayesian Estimation View: Probabilistic Optimality. }

By treating $\text{SW}_t^{(d)}$ as the likelihood from recent observations and $\text{EMA}_t^{(d)}$ as the prior estimate, the fused estimator can be interpreted as a posterior expectation:

\begin{equation}
\hat{w}_t^{(d)} = \frac{\tau^2}{\sigma^2 + \tau^2} \cdot \text{SW}_t^{(d)} + \frac{\sigma^2}{\sigma^2 + \tau^2} \cdot \text{EMA}_t^{(d)}
\end{equation}

\noindent where $\sigma^2$ and $\tau^2$ denote the variances of the short-term and long-term estimators, respectively. This justifies $\lambda$ as a data-dependent confidence weight, supporting the optimality of the fusion under uncertainty.

\subsubsection{2. Kalman Filtering Approximation: Recursive Preference Tracking. }

The update rule resembles a simplified Kalman filter:

\begin{equation}
\hat{w}_t^{(d)} = \hat{w}_{t-1}^{(d)} + K_t \cdot (p_t^{(d)} - \hat{w}_{t-1}^{(d)})
\end{equation}

\noindent with gain $K_t$ computed as:

\begin{equation}
K_t = \frac{P_{t|t-1}}{P_{t|t-1} + R}
\end{equation}

\noindent where $P_{t|t-1}$ is the prior variance and $R$ is the observation noise variance. Setting $K_t \approx (1 - \beta)$ shows the correspondence to EMA. This analogy supports the recursive structure and temporal filtering behavior of our mechanism.

%\subsubsection{3. Bias-Variance Tradeoff: Robust Estimation. }

%From statistical learning theory, the expected error of the estimator is:

%\begin{equation}
%\mathbb{E}\left[ \left( \hat{w}_t^{(d)} - \mu_t \right)^2 \right] = \text{Bias}^2 + \text{Variance}
%\end{equation}

%\noindent Sliding window has low bias but high variance, while EMA has high bias but low variance. Their convex combination minimizes the total risk, providing a balanced and generalizable estimate across both volatile and stable conditions.

\subsubsection{3. Change Detection Signal: Behavioral Adaptation Trigger. }

The deviation between SW and EMA serves as a change indicator:

\begin{equation}
\Delta_t^{(d)} = \left| \text{SW}_t^{(d)} - \text{EMA}_t^{(d)} \right|
\end{equation}

\noindent To normalize the change magnitude, we define a detection score:

\begin{equation}
C_t^{(d)} = \frac{\Delta_t^{(d)}}{\epsilon + \sqrt{\text{Var}(\text{SW}) + \text{Var}(\text{EMA})}}
\end{equation}

\noindent When $C_t^{(d)}$ exceeds a predefined threshold $\delta$, the system can trigger prompt rewriting, memory graph restructuring, or strategy modulation. This confirms the mechanism's role as an interpretable and actionable controller for preference-aware behavior.

This mechanism addresses the core problem proposed in this paper: how to dynamically update user preference memory within LLM agents and accordingly adjust their responses in real time.

\section{Experiment}

\subsection{Setup}

\subsubsection{Dataset and Evaluation Metrics. }To evaluate whether our preference update mechanism can effectively guide LLMs to generate user-aligned responses in long-term multi-turn dialogue scenarios, we adopt the LoCoMo dataset \cite{LoCoMoDataset} following previous related work \cite{Amem,memorybank}. LoCoMo is specifically designed to assess the memory and consistency capabilities of LLM-based agents in extended multi-session interactions. Key characteristics of the dataset include 50 dialogues, each with an average of 300 turns, spanning up to 35 distinct sessions and approximately 9,000 tokens per dialogue. We choose three types of task in it:
\begin{itemize}
    \item \textbf{Single-hop questions (SH.)}: answerable within a single session (2,705 pairs).
    \item \textbf{Multi-hop questions (MH.)}: requiring cross-session information aggregation (1,104 pairs).
    \item \textbf{Temporal reasoning (T.)}: testing understanding of time-sensitive information (1,547 pairs).
    %\item \textbf{Adversarial questions (A.)}: assessing the model’s ability to detect unanswerable cases (1,871 pairs).
\end{itemize}

LoCoMo emphasizes long-range contextual coherence across sessions, making it a robust benchmark for evaluating LLMs’ ability to handle memory-dependent reasoning and maintain response consistency in long-term interactions.

We employ two primary metrics to comprehensively assess model performance under different memory settings: \textbf{(1) F1 Score}: Measures the harmonic mean of precision and recall between the generated and reference answers, capturing semantic accuracy and completeness; \textbf{(2) BLEU-1 Score}: Evaluates the surface quality and fluency of generated responses via unigram overlap with the gold standard.
These metrics jointly assess the effectiveness of our mechanism in enhancing user-aligned generation in long-context conversational settings.

\subsubsection{Baselines. }As our work specifically focuses on preference memory update mechanisms rather than proposing a complete memory framework, we evaluate the effectiveness of our approach by integrating it into five representative long-term memory methods and conducting before-and-after comparisons. The selected baselines include: \textbf{ReadAgent (RA.)} \cite{readingagent}, \textbf{MemoryBank (MB.)} \cite{memorybank}, \textbf{MemGPT (MG.)} \cite{memgpt}, and \textbf{A-MEM (AM.)} \cite{Amem}, all of which are currently very mainstream memory frameworks. For each method, we augment its original architecture by appending our preference update module, without modifying its internal memory operations or update logic. Importantly, our mechanism is fully compatible and modular, operating independently of each baseline’s native update strategy. The only difference between the original and enhanced versions lies in the presence of our preference-aware update component, ensuring that any observed performance gains can be attributed solely to our proposed mechanism.

\subsubsection{Implementation Details. }In our experiments, we utilize three families of large language models with varying scales—Qwen 2.5-1.5B / 3B \cite{qwen2}, LLaMA-7B / 30B \cite{llama}, and LLaMA 3.2-1.5B / 3B \cite{llama}—as the base QA models. These diverse model types and sizes allow for a more comprehensive evaluation of the robustness and generalizability of our proposed mechanism. All models are deployed locally via Ollama. For our preference signal extraction, we employ the following pretrained models for each corresponding dimension: RoBERTa encoder with a multi-class classification head (Tone Style); Open-source pretrained SKEP \cite{skep} model (Emotional Tone); Knowledge tuples extracted via OpenNRE \cite{opennre}, representing structured semantic units (Information Density). To ensure fair comparisons and experimental reliability, we apply identical configurations of our preference module across all baseline memory systems. No modifications are made to their original architectures or reasoning logic, except for minimal adaptations to accommodate preference integration. During inference, each model receives only the input question and its respective historical memory. The final preference prompt—generated from the computed vector—is appended to the original input prompt of each method, providing explicit control signals to guide response generation.

\subsection{Main Results and Analysis}

Each result represents the average over three independent runs with different random seeds. We conducted paired t-tests among baselines. Results marked with $\textbf{*}$ indicate statistically significant improvements (p $<$ 0.05). $\dagger$ indicates the model is equipped with our proposed Preference-Aware Memory Update (PAMU) mechanism. The format of all results is \textbf{Before Augment / After Augment}. 

\subsubsection{Comparison Analysis. }As shown in Table \ref{tab:mh.}, \ref{tab:sh.} and  \ref{tab:t.}, our method was evaluated on three representative tasks. For both the single-hop and multi-hop reasoning tasks, all baselines equipped with PAMU demonstrated significant improvements in response quality, while maintaining or slightly improving accuracy. This highlights the effectiveness and generalizability of PAMU in enhancing generation without compromising correctness. Notably, in the temporal reasoning task, PAMU led to substantial gains in both accuracy and response quality, indicating its ability not only to detect short-term preference shifts but also to effectively update long-term user trends.

\begin{table}[h]
    \centering
  \begin{tabular}{ccccc}
    \toprule  
   \multicolumn{3}{c}{ \multirow{2}*{\textbf{Methods}}}  %&\multirow{2}*{\textbf{Baselines}}
    
    &\multicolumn{2}{c}{ \textbf{Single-Hop}}\\

    \cline{4-5}
    &&&\multicolumn{1}{c}{ \textbf{F1}}
    &\multicolumn{1}{c}{ \textbf{BLUE-1}}
    \\

       \toprule
  \multirow{8}*{\rotatebox{90}{\textbf{Qwen 2.5}}}&\multirow{4}*{\rotatebox{90}{\textbf{1.5B}}}  
  &RA. / RA.$\dagger$  & 6.54 / \textbf{8.27} &4.87 / \textbf{8.97*} \\
  &&MB. / MB.$\dagger$ & 11.14 / \textbf{12.34} & 8.24 / \textbf{10.57*} \\
  &&MG. / MG.$\dagger$ & 10.43 / \textbf{10.49}  & 7.54 / \textbf{11.46*}  \\
  &&AM. / AM.$\dagger$ & 17.24 / \textbf{17.93}   & 11.35 / \textbf{15.73*}   \\

  \cline{2-5}
  &\multirow{4}*{\rotatebox{90}{\textbf{3B}}}
  &RA. / RA.$\dagger$  & 3.23 / 3.23  &2.89 / \textbf{4.23*}  \\
  &&MB. / MB.$\dagger$ & 3.54 / \textbf{3.87}   & 3.39 /  \textbf{7.35*}  \\
  &&MG. / MG.$\dagger$ & 5.07 / \textbf{5.24}   &4.28 /  \textbf{8.65*}  \\
  &&AM. / AM.$\dagger$ & 12.52 / \textbf{13.23}   & 9.24 /  \textbf{13.24*}  \\
  
\toprule

\end{tabular}

 \caption{Experimental results on single-hop tasks using Qwen 2.5-1.5B/3B models.  }
\label{tab:sh.}
\end{table}

\begin{table}[h]
    \centering
  \begin{tabular}{ccccc}
    \toprule  
   \multicolumn{3}{c}{ \multirow{2}*{\textbf{Methods}}}  %&\multirow{2}*{\textbf{Baselines}}
    
    &\multicolumn{2}{c}{ \textbf{Multi-Hop}}\\

    \cline{4-5}
    &&&\multicolumn{1}{c}{ \textbf{F1}}
    &\multicolumn{1}{c}{ \textbf{BLUE-1}}
    \\

       \toprule
  \multirow{8}*{\rotatebox{90}{\textbf{LLaMA 3.2}}}&\multirow{4}*{\rotatebox{90}{\textbf{1.5B}}}  
  &RA. / RA.$\dagger$  &2.45 / \textbf{2.98}  &2.67 / \textbf{5.34*}\\
  &&MB. / MB.$\dagger$ & 7.61 / \textbf{6.03} & 6.56 /  \textbf{9.23*}\\
  &&MG. / MG.$\dagger$ & 5.23 / \textbf{6.78} & 5.14 / \textbf{10.87*}\\
  &&AM. / AM.$\dagger$ & 16.57 / \textbf{17.02} & 11.24 / \textbf{19.23*} \\

  \cline{2-5}
  &\multirow{4}*{\rotatebox{90}{\textbf{3B}}}
  &RA. / RA.$\dagger$  & 3.05 / \textbf{3.67} &2.67 / \textbf{5.45*}\\
  &&MB. / MB.$\dagger$ & 3.56 / 3.56 & 3.02 / \textbf{7.65*} \\
  &&MG. / MG.$\dagger$ & 3.02 / 3.02 & 2.95 / \textbf{6.34*}\\
  &&AM. / AM.$\dagger$ & 19.35 / \textbf{20.14} & 13.27 / \textbf{23.14*} \\
  
\toprule

\end{tabular}

 \caption{Experimental results on multi-hop tasks using LLaMA 3.2-1.5B/3B models.  }
\label{tab:mh.}
\end{table}

\begin{table}[h]
    \centering
  \begin{tabular}{ccccc}
    \toprule  
   \multicolumn{3}{c}{ \multirow{2}*{\textbf{Methods}}}  %&\multirow{2}*{\textbf{Baselines}}
    
    &\multicolumn{2}{c}{ \textbf{Temporal Reasoning}}\\

    \cline{4-5}
    &&&\multicolumn{1}{c}{ \textbf{F1}}
    &\multicolumn{1}{c}{ \textbf{BLUE-1}}
    \\

       \toprule
  \multirow{8}*{\rotatebox{90}{\textbf{LLaMA}}}&\multirow{4}*{\rotatebox{90}{\textbf{7B}}}  
  &RA. / RA.$\dagger$  &12.24 / \textbf{15.45*}  &11.17 / \textbf{15.67*}\\
  &&MB. / MB.$\dagger$ & 14.56 / \textbf{19.76*} & 11.95 / \textbf{17.24*} \\
  &&MG. / MG.$\dagger$ & 11.14 / \textbf{17.54*} &8.24 / \textbf{15.57*} \\
  &&AM. / AM.$\dagger$ &  17.55 / \textbf{23.23*}& 14.67 / \textbf{21.46*} \\

  \cline{2-5}
  &\multirow{4}*{\rotatebox{90}{\textbf{30B}}}
  &RA. / RA.$\dagger$  & 5.57 / \textbf{7.67*} &5.22 / \textbf{7.43*}\\
  &&MB. / MB.$\dagger$ & 4.77 / \textbf{8.98*} &  4.87 / \textbf{7.34*}\\
  &&MG. / MG.$\dagger$ & 5.64 / \textbf{9.95*} & 5.53 / \textbf{8.24*}\\
  &&AM. / AM.$\dagger$ & 12.54 / \textbf{19.87*} & 11.85 / \textbf{18.23*} \\
  
\toprule

\end{tabular}

 \caption{Experimental results on temporal reasoning tasks using LLaMA-7B/30B models.  }
\label{tab:t.}
\end{table}

\subsubsection{Ablation Study. }

\begin{table}[h]
    \centering
  
  \begin{tabular}{ccccc}
    \toprule  
   \multirow{1}*{ \textbf{Methods}}  
    &\multicolumn{1}{c}{ \textbf{RA.$\dagger$}}

    &\multirow{1}*{\textbf{MB.$\dagger$}}
    &\multirow{1}*{\textbf{MG.$\dagger$}}
    &\multirow{1}*{\textbf{AM.$\dagger$}}
    
 \\
  \toprule
  
  w/o. SW & 11.24& 12.03 &10.07& 15.36\\

  w/o. EMA &11.35&12.47&10.78&14.05\\
  Equal Fusion &13.56&16.45&15.43&20.34\\
  w/o Detection &12.34&13.28&12.24&16.24\\
  w/o Prompt &11.13&12.25&9.37&15.45\\
  Single Pref &12.21&16.78&14.23&18.95\\
  Static Pref &12.34&16.21&13.24&19.47\\

\textbf{Full}&\textbf{15.56} &\textbf{18.50}& \textbf{16.56} &\textbf{22.35}\\
    
    \toprule
\end{tabular}  
 \caption{Ablation Study. We select the experimental results of temporal reasoning using LLaMA 7B and took the average of F1 and BLUE-1. }
\label{tab:ablation}
\end{table}

\begin{table*}[h]
    \centering
%\scriptsize
  
  \begin{tabular}{cccccc}
    \toprule  
   \multirow{1}*{ \textbf{Turn}}  
    &\multicolumn{1}{c}{ \textbf{Tone.}}

    &\multirow{1}*{\textbf{Length}}
    &\multirow{1}*{\textbf{Emotion}}
    &\multirow{1}*{\textbf{Density}}
    &\multirow{1}*{\textbf{Formality}}
    
 \\
      % \toprule
  %Mistral-7B  &0.17 &3.25  \\
  %Alpaca-7B& 5.88&5.81  \\
  \toprule
  
  1& 	(Humor, 0.92)& 0.18 &(Joy, 0.85) &0.20&	0.15\\%&&AM. &18.23 &11.94 &24.32 &19.74  &23.63 &19.23 &46.00 &43.26 & &\\

  2&(Humor, 0.93)&0.16&	(Joy, 0.86)&0.22&	0.17\\
  3 &(Neutral, 0.72)&0.45&	(Neutral, 0.70)&0.55&	0.48\\
4&(Serious, 0.89)&0.71&	(Focused, 0.91)&	0.78&	0.80\\
5&(Serious, 0.95)&0.69&	(Neutral, 0.88)&	0.82&0.85\\
    
    \toprule
\end{tabular}

 \caption{Data extracted from the designed dialogues using the preference extractor in PAMU.  }
\label{tab:Preference_exactor}
\end{table*}

To systematically evaluate the individual contributions of each component in our proposed preference-aware memory update mechanism, we conduct a comprehensive set of ablation studies. The details of each ablation and its corresponding replacement are as follows:

\begin{itemize}

\item Sliding Window Average (w/o SW): Captures short-term preference shifts. Ablation removes SW, leaving only EMA to simulate lack of short-term responsiveness.
\item Exponential Moving Average (w/o EMA): Models long-term preference trends. Removing EMA isolates the effect of losing long-term stability.
\item Fusion Mechanism (Equal Fusion): The original model learns a dynamic weight $\lambda$ to fuse SW and EMA. Ablation fixes $\lambda$ = 0.5, disabling adaptive balancing.
\item Preference Change Detection (w/o Detection): Removes the divergence-based change signal, preventing prompt/memory adaptation and reverting to static generation templates.
\item Prompt Injection (w/o Prompt): Eliminates explicit preference prompts, providing only raw user input to test generation without direct conditioning.

\item Multi-Dimensional Preference Modeling (Single Pref): Reduces the 5D preference vector (tone, length, emotion, density, formality) to a single feature (e.g., length) to assess the benefit of multi-dimensional modeling.
\item Dynamic vs. Static Preference Modeling (Static Pref): Replaces dynamically updated preference with a fixed vector averaged over the first five turns, simulating static memory systems.

\end{itemize}

All ablations are conducted under consistent training settings, model architectures, and evaluation protocols to ensure causal interpretability. Experimental results are shown in Table \ref{tab:ablation}, it can be seen that each module plays an essential and non-redundant role in maintaining consistency, personalization, and preference alignment throughout long-term interactions.

\section{Further Analysis}
To further demonstrate the interpretability and responsiveness of our Preference-Aware Memory Update (PAMU) mechanism, we design a controlled dialogue-based case study simulating a typical shift in user preference. %This experiment aims to evaluate whether PAMU can (1) accurately extract multi-dimensional preferences, (2) detect abrupt or gradual shifts, and (3) update the generation strategy accordingly.

Since our mechanism is subjective and there is no objective metric to evaluate the specific effectiveness of its components, we incorporate both GPT-4 automatic scoring and human judgment. The human evaluation was conducted by ten annotators with bachelor's degrees, completed over a two-week period. Annotators were instructed not to use any AI tools during the assessment to ensure manual, unbiased evaluation. Notably, none of the paper's authors participated in the evaluation phase, ensuring fairness and neutrality.

We design the following dialogue to evaluate whether the components in PAMU mechanism are effective.

\begin{tcolorbox}[colback=gray!5!white, colframe=gray!50!black, title=Example Dialogue]

\textbf{Turn 1 (User)}: Hey, tell me something and funny!

\textbf{Turn 2 (User)}: That's good! I like it short and fun.

\textbf{Turn 3 (User)}: Actually, I have a serious task now. Can you be more detailed?

\textbf{Turn 4 (User)}: I need a thorough explanation on quantum computing basics. 

\textbf{Assistant (with PAMU)}: Certainly. Quantum computing is based on quantum bits, or qubits... %which can exist in multiple states simultaneously...

\textbf{Assistant}: Sure! Here’s another fact to brighten you...

\textbf{Turn 5 (User)}: Please just give me clear facts.

\end{tcolorbox}

As shown above, the user initially demonstrates a clear preference for humorous and concise responses (Turns 1–2), but this preference abruptly shifts toward formal and information-dense content starting from Turn 3. PAMU captures this shift in real time by monitoring the divergence between the short-term (SW) and long-term (EMA) estimates for each preference dimension. At Turn 3, the preference change signal $C_t^{(d)}$ surpasses the predefined threshold $\delta$ in multiple dimensions (e.g., tone, length, density), triggering an immediate update to the fused preference vector $\hat{w}_t$ and rewriting of the prompt. Preference Vector Dynamics (Extracted) is shown in Table \ref{tab:Preference_exactor}.

To evaluate the utility of PAMU’s dynamic prompting, we compare model outputs with and without PAMU at Turn 4. Without PAMU, the model continues generating light, humorous content, misaligned with the user's updated intent. In contrast, the PAMU-augmented response accurately adapts in tone, density, and formality, reflecting a meaningful understanding of user behavior change, as shown in dialogue content. Additionally, results in Tables \ref{tab:compare1} and \ref{tab:compare2} further demonstrate the effectiveness of PAMU.

\begin{table}[h]
    \centering
%\scriptsize
  
  \begin{tabular}{cccc}
    \toprule  
   \multirow{1}*{ \textbf{Turn}}  
    &\multicolumn{1}{c}{ \textbf{Align(1-5)}}

    &\multirow{1}*{\textbf{Cons.}}
    &\multirow{1}*{\textbf{Response speed}}

 \\
      % \toprule
  %Mistral-7B  &0.17 &3.25  \\
  %Alpaca-7B& 5.88&5.81  \\
  \toprule
  
  w/o PAMU & 2.1/2.2 & \ding{55}/\ding{55} &Two-round delay \\%&&AM. &18.23 &11.94 &24.32 &19.74  &23.63 &19.23 &46.00 &43.26 & &\\

  with PAMU&4.8/4.5 &\ding{51}/\ding{51}&	Real-time (Turn 3)\\

    \toprule
\end{tabular}

 \caption{Comparison results, scoring results are in the format of (GPT/Human). Cons. represents consistency. }
\label{tab:compare1}
\end{table}

\begin{table}[h]
    \centering
%\scriptsize
  
  \begin{tabular}{ccc}
    \toprule  
   \multirow{1}*{ \textbf{Methods}}  
    &\multicolumn{1}{c}{ \textbf{w/o PAMU}}

    &\multirow{1}*{\textbf{with PAMU}}

 \\
      % \toprule
  %Mistral-7B  &0.17 &3.25  \\
  %Alpaca-7B& 5.88&5.81  \\
  \toprule
  
  Style Consistency (\%)& 37/35 &  92/94 \\%&&AM. &18.23 &11.94 &24.32 &19.74  &23.63 &19.23 &46.00 &43.26 & &\\

  Preference detection (\%)&48/45 &97/95\\

    \toprule
\end{tabular}

 \caption{Comparison results, scoring results are in the format of (GPT/Human). }
\label{tab:compare2}
\end{table}

This case study confirms that PAMU can dynamically track evolving user preferences, detect both abrupt and gradual changes, and trigger appropriate generation adaptations, leading to more personalized, user-aligned interactions. %It also highlights the interpretability of PAMU’s internal dynamics, a key criterion for trustworthy human-LLM interaction systems.

\section{Conclusion}

We propose a Preference-Aware Memory Update Mechanism to address the limitations of existing memory systems in tracking evolving user preferences. By combining sliding window and exponential moving averages, our method captures both short-term dynamics and long-term trends. A formalized change detection signal—based on their divergence—triggers memory updates, enabling interpretable and adaptive preference-aware behavior.

%In this paper, we proposed a Preference-Aware Memory Update Mechanism to address the limitations of existing memory systems in adapting to evolving user preferences. By combining sliding window averages (SW) with exponential moving averages (EMA), our approach constructs a preference representation that is both sensitive to short-term fluctuations and robust to long-term trends. Furthermore, we introduced a formal change detection signal that triggers memory updates based on the deviation between short- and long-term estimates, enabling interpretable and controllable adaptation. %Experimental results demonstrate that our mechanism significantly enhances the agent’s ability to track user evolution, offering a novel and effective pathway for building more adaptive memory-augmented LLM agents.

\bibliography{main}

\begin{thebibliography}{24}
\providecommand{\natexlab}[1]{#1}

\bibitem[{DeepSeek-AI(2025)}]{deepseek-r1}
DeepSeek-AI. 2025.
\newblock DeepSeek-R1: Incentivizing Reasoning Capability in LLMs via Reinforcement Learning.
\newblock arXiv:2501.12948.

\bibitem[{Gu et~al.(2024)Gu, Jiang, Shi, Tan, Zhai, Xu, Li, Shen, Ma, Liu et~al.}]{LLMsurvey2}
Gu, J.; Jiang, X.; Shi, Z.; Tan, H.; Zhai, X.; Xu, C.; Li, W.; Shen, Y.; Ma, S.; Liu, H.; et~al. 2024.
\newblock A survey on llm-as-a-judge.
\newblock \emph{arXiv preprint arXiv:2411.15594}.

\bibitem[{Guo et~al.(2024)Guo, Chen, Wang, Chang, Pei, Chawla, Wiest, and Zhang}]{agentsurvey3}
Guo, T.; Chen, X.; Wang, Y.; Chang, R.; Pei, S.; Chawla, N.~V.; Wiest, O.; and Zhang, X. 2024.
\newblock Large language model based multi-agents: A survey of progress and challenges.
\newblock \emph{arXiv preprint arXiv:2402.01680}.

\bibitem[{Han et~al.(2019)Han, Gao, Yao, Ye, Liu, and Sun}]{opennre}
Han, X.; Gao, T.; Yao, Y.; Ye, D.; Liu, Z.; and Sun, M. 2019.
\newblock OpenNRE: An open and extensible toolkit for neural relation extraction.
\newblock \emph{arXiv preprint arXiv:1909.13078}.

\bibitem[{Huang et~al.(2024{\natexlab{a}})Huang, Liu, Chen, Wang, Wang, Lian, Wang, Tang, and Chen}]{agentsurvey}
Huang, X.; Liu, W.; Chen, X.; Wang, X.; Wang, H.; Lian, D.; Wang, Y.; Tang, R.; and Chen, E. 2024{\natexlab{a}}.
\newblock Understanding the planning of LLM agents: A survey.
\newblock \emph{arXiv preprint arXiv:2402.02716}.

\bibitem[{Huang et~al.(2024{\natexlab{b}})Huang, Liu, Chen, Wang, Wang, Lian, Wang, Tang, and Chen}]{LLMsurvey6}
Huang, X.; Liu, W.; Chen, X.; Wang, X.; Wang, H.; Lian, D.; Wang, Y.; Tang, R.; and Chen, E. 2024{\natexlab{b}}.
\newblock Understanding the planning of LLM agents: A survey.
\newblock \emph{arXiv preprint arXiv:2402.02716}.

\bibitem[{Jin et~al.(2024)Jin, Huang, Cai, Yan, Li, and Chen}]{agentsurvey4}
Jin, H.; Huang, L.; Cai, H.; Yan, J.; Li, B.; and Chen, H. 2024.
\newblock From llms to llm-based agents for software engineering: A survey of current, challenges and future.
\newblock \emph{arXiv preprint arXiv:2408.02479}.

\bibitem[{Lee et~al.(2024)Lee, Chen, Furuta, Canny, and Fischer}]{readingagent}
Lee, K.-H.; Chen, X.; Furuta, H.; Canny, J.; and Fischer, I. 2024.
\newblock A Human-Inspired Reading Agent with Gist Memory of Very Long Contexts.
\newblock arXiv:2402.09727.

\bibitem[{Li et~al.(2024)Li, Wen, Wang, Li, Yuan, Liu, Liu, Xu, Wang, Sun et~al.}]{agentsurvey2}
Li, Y.; Wen, H.; Wang, W.; Li, X.; Yuan, Y.; Liu, G.; Liu, J.; Xu, W.; Wang, X.; Sun, Y.; et~al. 2024.
\newblock Personal llm agents: Insights and survey about the capability, efficiency and security.
\newblock \emph{arXiv preprint arXiv:2401.05459}.

\bibitem[{Maharana et~al.(2024)Maharana, Lee, Tulyakov, Bansal, Barbieri, and Fang}]{LoCoMoDataset}
Maharana, A.; Lee, D.-H.; Tulyakov, S.; Bansal, M.; Barbieri, F.; and Fang, Y. 2024.
\newblock Evaluating very long-term conversational memory of llm agents.
\newblock \emph{arXiv preprint arXiv:2402.17753}.

\bibitem[{Packer et~al.(2023)Packer, Fang, Patil, Lin, Wooders, and Gonzalez}]{memgpt}
Packer, C.; Fang, V.; Patil, S.; Lin, K.; Wooders, S.; and Gonzalez, J. 2023.
\newblock MemGPT: Towards LLMs as Operating Systems.

\bibitem[{Salama et~al.(2025)Salama, Cai, Yuan, Currey, Sunkara, Zhang, and Benajiba}]{meminsight}
Salama, R.; Cai, J.; Yuan, M.; Currey, A.; Sunkara, M.; Zhang, Y.; and Benajiba, Y. 2025.
\newblock MemInsight: Autonomous Memory Augmentation for LLM Agents.
\newblock \emph{arXiv preprint arXiv:2503.21760}.

\bibitem[{Sun et~al.(2025)Sun, Bian, Zeng, Rao, Xu, Mei, and Gou}]{datasetagent}
Sun, H.; Bian, H.; Zeng, S.; Rao, Y.; Xu, X.; Mei, L.; and Gou, J. 2025.
\newblock DatasetAgent: A Novel Multi-Agent System for Auto-Constructing Datasets from Real-World Images.
\newblock arXiv:2507.08648.

\bibitem[{Sun and Zeng(2025)}]{INoT}
Sun, H.; and Zeng, S. 2025.
\newblock Introspection of Thought Helps AI Agents.
\newblock arXiv:2507.08664.

\bibitem[{Tian et~al.(2020)Tian, Gao, Xiao, Liu, He, Wu, Wang, and Wu}]{skep}
Tian, H.; Gao, C.; Xiao, X.; Liu, H.; He, B.; Wu, H.; Wang, H.; and Wu, F. 2020.
\newblock SKEP: Sentiment Knowledge Enhanced Pre-training for Sentiment Analysis.
\newblock arXiv:2005.05635.

\bibitem[{Touvron et~al.(2023)Touvron, Lavril, Izacard, Martinet, Lachaux, Lacroix, Rozière, Goyal, Hambro, Azhar, Rodriguez, Joulin, Grave, and Lample}]{llama}
Touvron, H.; Lavril, T.; Izacard, G.; Martinet, X.; Lachaux, M.-A.; Lacroix, T.; Rozière, B.; Goyal, N.; Hambro, E.; Azhar, F.; Rodriguez, A.; Joulin, A.; Grave, E.; and Lample, G. 2023.
\newblock LLaMA: Open and Efficient Foundation Language Models.
\newblock arXiv:2302.13971.

\bibitem[{Wang et~al.(2023)Wang, Liang, Yang, Huang, Wu, Wu, Lu, Ma, and Li}]{SCM}
Wang, B.; Liang, X.; Yang, J.; Huang, H.; Wu, S.; Wu, P.; Lu, L.; Ma, Z.; and Li, Z. 2023.
\newblock Enhancing large language model with self-controlled memory framework.
\newblock \emph{arXiv preprint arXiv:2304.13343}.

\bibitem[{Wu et~al.(2025)Wu, Yang, Zhan, Yuan, Chao, and Wong}]{LLMsurvey5}
Wu, J.; Yang, S.; Zhan, R.; Yuan, Y.; Chao, L.~S.; and Wong, D.~F. 2025.
\newblock A survey on LLM-generated text detection: Necessity, methods, and future directions.
\newblock \emph{Computational Linguistics}, 1--66.

\bibitem[{Xu et~al.(2025)Xu, Liang, Mei, Gao, Tan, and Zhang}]{Amem}
Xu, W.; Liang, Z.; Mei, K.; Gao, H.; Tan, J.; and Zhang, Y. 2025.
\newblock A-mem: Agentic memory for llm agents.
\newblock \emph{arXiv preprint arXiv:2502.12110}.

\bibitem[{Yang et~al.(2024)Yang, Yang, Hui, Zheng, Yu, Zhou, Li, Li, Liu, Huang, Dong, Wei, Lin, Tang, Wang, Yang, Tu, Zhang, Ma, Xu, Zhou, Bai, He, Lin, Dang, Lu, Chen, Yang, Li, Xue, Ni, Zhang, Wang, Peng, Men, Gao, Lin, Wang, Bai, Tan, Zhu, Li, Liu, Ge, Deng, Zhou, Ren, Zhang, Wei, Ren, Fan, Yao, Zhang, Wan, Chu, Liu, Cui, Zhang, and Fan}]{qwen2}
Yang, A.; Yang, B.; Hui, B.; Zheng, B.; Yu, B.; Zhou, C.; Li, C.; Li, C.; Liu, D.; Huang, F.; Dong, G.; Wei, H.; Lin, H.; Tang, J.; Wang, J.; Yang, J.; Tu, J.; Zhang, J.; Ma, J.; Xu, J.; Zhou, J.; Bai, J.; He, J.; Lin, J.; Dang, K.; Lu, K.; Chen, K.; Yang, K.; Li, M.; Xue, M.; Ni, N.; Zhang, P.; Wang, P.; Peng, R.; Men, R.; Gao, R.; Lin, R.; Wang, S.; Bai, S.; Tan, S.; Zhu, T.; Li, T.; Liu, T.; Ge, W.; Deng, X.; Zhou, X.; Ren, X.; Zhang, X.; Wei, X.; Ren, X.; Fan, Y.; Yao, Y.; Zhang, Y.; Wan, Y.; Chu, Y.; Liu, Y.; Cui, Z.; Zhang, Z.; and Fan, Z. 2024.
\newblock Qwen2 Technical Report.
\newblock \emph{arXiv preprint arXiv:2407.10671}.

\bibitem[{Yao et~al.(2024)Yao, Duan, Xu, Cai, Sun, and Zhang}]{LLMsurvey1}
Yao, Y.; Duan, J.; Xu, K.; Cai, Y.; Sun, Z.; and Zhang, Y. 2024.
\newblock A survey on large language model (llm) security and privacy: The good, the bad, and the ugly.
\newblock \emph{High-Confidence Computing}, 100211.

\bibitem[{Zhang et~al.(2024{\natexlab{a}})Zhang, Bo, Ma, Li, Chen, Dai, Zhu, Dong, and Wen}]{memorysurvey}
Zhang, Z.; Bo, X.; Ma, C.; Li, R.; Chen, X.; Dai, Q.; Zhu, J.; Dong, Z.; and Wen, J.-R. 2024{\natexlab{a}}.
\newblock A survey on the memory mechanism of large language model based agents.
\newblock \emph{arXiv preprint arXiv:2404.13501}.

\bibitem[{Zhang et~al.(2024{\natexlab{b}})Zhang, Bo, Ma, Li, Chen, Dai, Zhu, Dong, and Wen}]{memorysurvey1}
Zhang, Z.; Bo, X.; Ma, C.; Li, R.; Chen, X.; Dai, Q.; Zhu, J.; Dong, Z.; and Wen, J.-R. 2024{\natexlab{b}}.
\newblock A survey on the memory mechanism of large language model based agents.
\newblock \emph{arXiv preprint arXiv:2404.13501}.

\bibitem[{Zhong et~al.(2024)Zhong, Guo, Gao, Ye, and Wang}]{memorybank}
Zhong, W.; Guo, L.; Gao, Q.; Ye, H.; and Wang, Y. 2024.
\newblock Memorybank: Enhancing large language models with long-term memory.
\newblock In \emph{Proceedings of the AAAI Conference on Artificial Intelligence}, volume~38, 19724--19731.

\end{thebibliography}

\end{document}